\documentclass[conference]{IEEEtran}
\IEEEoverridecommandlockouts
\usepackage[utf8]{inputenc}
\usepackage{csquotes}
\usepackage{caption}
\usepackage{cite}
\usepackage{amsmath,amssymb,amsfonts}
\usepackage{algorithmic}
\usepackage{graphicx}
\usepackage{textcomp}
\usepackage{xcolor}
\usepackage{hhline}	
\usepackage{subfigure}
\usepackage{listings}
\def\BibTeX{{\rm B\kern-.05em{\sc i\kern-.025em b}\kern-.08em
		T\kern-.1667em\lower.7ex\hbox{E}\kern-.125emX}}
\usepackage[]{todonotes}

\begin{document}
	\title{Reliable validation of\\Reinforcement Learning Benchmarks}
	
	\author{\IEEEauthorblockN{Matthias Müller-Brockhausen, Aske Plaat, Mike Preuss}
		\IEEEauthorblockA{\textit{Leiden Institute of Advanced Computer Science (LIACS)} \\
			\textit{Leiden University}\\
			The Netherlands \\
			\{m.f.t.muller-brockhausen, a.plaat, m.preuss\}@liacs.leidenuniv.nl}
	}
	\maketitle
	
	\begin{abstract}
		Reinforcement Learning (RL) is one of the most dynamic research areas in Game AI and AI as a whole, and a wide variety of games are used as its prominent test problems.
		However, it is subject to the replicability crisis that currently affects most algorithmic AI research.
		Benchmarking in Reinforcement Learning could be improved through verifiable results. There are numerous benchmark environments whose scores are used to compare different algorithms, such as Atari. Nevertheless, reviewers must trust that figures represent truthful values, as it is difficult to reproduce an exact training curve. 
		We propose improving this situation by providing access to the original experimental data to validate study results.
		To that end, we rely on the concept of \textit{minimal} traces. These allow re-simulation of action sequences in deterministic RL environments and, in turn, enable reviewers to verify, re-use, and manually inspect experimental results without needing large compute clusters.
		It also permits validation of presented reward graphs, an inspection of individual episodes, and re-use of result data (baselines) for proper comparison in follow-up papers. We offer plug-and-play code that works with Gym so that our measures fit well in the existing RL and reproducibility eco-system.
		Our approach is freely available, easy to use, and adds minimal overhead, as \textit{minimal} traces allow a data compression ratio of up to $\approx 10^4:1$ (94~GB to 8~MB for Atari Pong) compared to a regular MDP trace used in offline RL datasets. The paper presents proof-of-concept results for a variety of games.
	\end{abstract}
	
	
	\section{Introduction}
	
	Reproducibility is a key component of peer-reviewed science. Reviewers are supposed to read, understand, and ideally be able to reproduce an experiment to ensure its factual correctness. It touches not only computer science, but any science, as without easy reproducibility, fraud is difficult to detect\cite{national2019reproducibility}. Especially for benchmarks and competitions, where fraudulent submissions potentially poison the rankings of a leaderboard, it is important to have tools for validation.
	
	Benchmarking AI algorithms has become increasingly important and is now a driving force behind algorithm development. In Game AI, competitions have been an important part of scientific conferences for a long time already, and especially game problem benchmarks are currently more and more spreading out to core AI conferences, e.g., with the MineRL competition at NeurIPS~\cite{guss2019the, guss2021the}. Whereas the overall aims of algorithm development are often to improve generality and especially sample efficiency, the employed methods are still relatively slow and thus need very long runs, thereby making reproducibility difficult.
	
	In theory, computers are excellent for reproducibility. One can run the same code, bit for bit, on many different machines. This may be simplified down to issuing a single command, based on technologies such as Docker\cite{boettiger2015introduction}.
	However, non-deterministic methods (e.g., evolutionary algorithms\cite{lopez2021reproducibility}, deep neural networks\cite{liu2020replicability}) hamper the reproducibility of experiments.
	Another growing problem is the availability of computing resources that would be needed to replicate results. 
	The tremendous successes of AlphaStar~\cite{AlphaStar} and Dota~2\cite{Dota2} are prominent examples.
	The large computing clusters they relied on are unavailable to most researchers for running any type of replication experiment.
	Furthermore, it becomes increasingly important to also consider sustainability issues, as the big cluster experiments are energy inefficient.
	Such considerations have been voiced, e.g., for Natural Language Processing (NLP)\cite{strubell2019energy} or complex Games as Go (AlphaZero)\cite{schwartz2020green}. It would thus make more sense to avoid re-computing everything but to improve the inspection of existing log data.
	Other issues that stand in the way of exact replication include insufficient reporting\cite{henderson2018deep} or not open-sourcing code\cite{pineau2021improving}.
	
	If replication itself is unavailable for some experiments, the next best thing could be verifiability, namely the ability to inspect, check, and replay parts of the experiment. However, this is also difficult even in terms of handling the huge amounts of data that are produced during the big experiments. In order to achieve this, we would need some way of highly compressing this data, which instantly points us to the concept of \emph{minimal traces}.

	The research question we are going to tackle in this work is thus: \emph{How may a researcher verify the reinforcement learning experiment of other researchers, especially the display of results in figures and tables, based on minimal traces?}
	\medskip
	
	\noindent We offer the following contributions:
	\begin{itemize}
		\item{We explain how minimal traces (Section \ref{sec:min-trace}) allow reproducible verification of results such as benchmark leaderboards (Section \ref{sec:method}). Moreover, we empirically show that they enable a compression ratio of up to $10^4:1$ for offline RL datasets (Section \ref{sec:compression})}
		\item{We provide Plug-n-Play code\cite{code} to collect minimal traces that integrate with the RL-Ecosystem (Gym\cite{gym})}
		\item{We provide an agenda for further research on how to obtain verifiable RL experiment results using minimal traces (Section \ref{sec:agenda})}
		
	\end{itemize}
	
	Although there are many factors at play with reproducibility, our work solely focuses on methodological improvements for reinforcement learning research. 
	After briefly introducing the concept of minimal traces (section \ref{sec:min-trace}), we first 
	look at suggestions that have already been made for improving reproducibility or experimental methodology in section \ref{sec:relatedwork}. Based on that state, we find improvable points (section \ref{sec:method}) and suggest concrete, actionable steps (section \ref{sec:agenda}). We then outline how these steps can be applied in practice with the code that we provide (section \ref{sec:apply}). To enable compatibility, we ensured that it properly interfaces to the existing RL-Ecosystem.
	
	\section{Background: Minimal Traces}
	\label{sec:min-trace}
	
	For the following sections, the concept of minimal traces is important, thus we review its origins and known uses here. 
	
	Reinforcement learning optimizes sequential decision making processes, that are modeled as so called Markov decision processes (MDPs).
	An MDP consists of a tuple ($S$, $A$, $P_a(s,s')$, $R_a(s, s')$) (state space, action space, the probability to go from state s to $s'$, and reward for going from s to $s'$)\cite{mdp}.
	
	A \textit{trace}, also commonly referred to as an Episode within an RL-Environment\cite{plaat2021drl}, is a list of tuples that contain the start state $s_t$, the chosen action $a$, the received reward $r$, and the resulting state $s_{t+1}$. Traces are sufficient to train an RL Algorithm offline / off-policy, and they are also shared by related work as dataset-basis for training\cite{agarwal2020optimistic}.

	Staying true to the computer science reinforcement learning terminology drawing inspiration from psychology\cite{sutton2018introduction}, we found a fitting concept to our problem, namely \textit{minimal traces}. Their goal seems related: \enquote{Predicting the Past from Minimal Traces}\cite{werning2020predicting}.
	We want reviewers to reliably predict (verify) the past (experiment results) using minimal traces.
	To reduce the used space (minimal) of traces we assume that an MDP, given the same initial state $S_0$ and action sequence $\alpha$, will yield the exact same trace. In consequence, the probabilities of $P_a(s,s')$ need to be fixed based on an initial configuration $s_{init}$. Fixing these probabilities is commonly referred to as a deterministic MDP\cite{post2015simplex}.
	Deterministic MDPs reduce the required data for re-simulation of minimal traces to ($s_{init}$, $s_0$, $\alpha_{t_0} ... \alpha_{t_n}$).
	Minimal traces fit well into reinforcement learning problems as there the action set $A$ is usually smaller than the state space $S$. Hence it makes more sense to only save actions, if the observations can be re-constructed afterward. While the added re-simulation cost might seem impractical for verification purposes, our Experiments show that it is can requires less than 0.7\% of the original RL training time (Section \ref{sec:compression}).
	\section{Related Work}
	\label{sec:relatedwork}

	While the matters of reproducibility, replicability, and verifiability are relevant to all scientific fields\cite{national2019reproducibility}, we will focus on reinforcement learning here.
	In reinforcement learning, previous works suggest guidelines on how to design and report a well reproducible experiment\cite{henderson2018deep}.
	Conferences such as NeurIPS are moving towards implementing these guidelines and ask reviewers to fill in a questionnaire about reproducibility. This has led to more and more sharing of code, and researchers are encouraged to do so \cite{pineau2021improving}. Moreover, reviewers found it easier to judge submissions that included code.
	Most of the reviewer guidelines focus on the paper itself, which is the well-established scientific tradition that was practiced already when computers were not yet invented. However, many researchers in Computer Science now believe that for experimental works, we should go one step further and exploit its theoretically perfect and exact ability to verify factual correctness of reported results and submitted code / data (Section \ref{sec:method}). This so-called \enquote{Verification of Artifacts}\cite{lopez2021reproducibility} is not a new concept.
	For example, tools available to make policy training as reproducible as possible are readily available (e.g. Garage\cite{garage}). For experiments that are not using tools as Garage, there are also clear guidelines on how to properly compare to a baseline of an algorithm\cite{islam2017reproducibility}.
	Moreover, researchers have suggested to save the final values used in graphs to be able to verify the figures\cite{khetarpal2018re}. This theoretically works for any Figure. However, how can we be sure that the Figure is correct\cite{eisner2018reproducibility}? 
	
	Games are an interesting playground for RL-AI. GVGAI is a prominent example\cite{perez2016analyzing}, as it is used for competitions that benchmark individual algorithm submissions from both planning\cite{gaina20172016} and learning such as RL\cite{torrado2018deep}.
	For these competitions, the validity of results is guaranteed by means of execution of the submitted code by the event host. This is made possible through a pre-defined agent interface, that allows to interact with arbitrary games. In other reinforcement learning environments, we also have a pre-defined agent interface (see the center of Figure \ref{fig:rl}), but re-running policy training is not at all reproducible. Reproducibility is not guaranteed in GVGAI either, as the applied algorithms, such as MCTS\cite{browne2012survey}, include randomness that is not fixed. While GVGAI remedies this by means of multiple runs, research has shown that averaging over runs does not prevent inconsistencies in reproduction attempts\cite{henderson2018deep}.
	Whereas minimal traces do not alleviate the problem directly, they do lift the requirement to have to run the agent code oneself.
	
	\subsection{Why minimal traces?}
	\label{sec:method} 
	\begin{figure}
		\centering
		\includegraphics[width=0.5\textwidth]{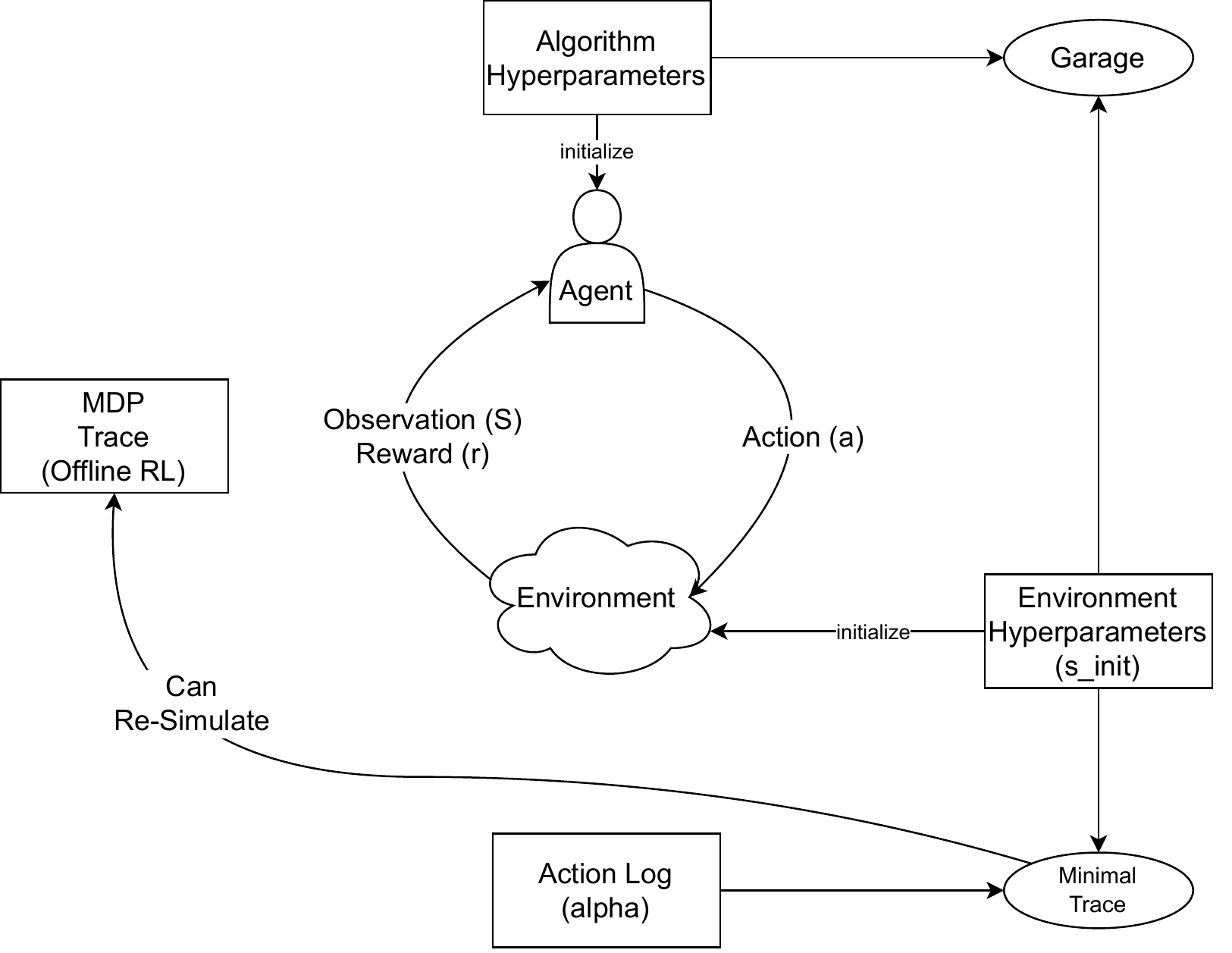}
		\caption{A visualization of the relationship between reinforcement learning training, offline RL datasets (traces), minimal traces Garage\cite{garage}, as well as the data collected for each.}
		\label{fig:rl}
	\end{figure} 
	Reproducibility in the RL ecosystem is an evolving matter. Environments usually behave deterministically if seeded, as, e.g., Cartpole or MountainCar in Gym\cite{gym}. Famous problems that did not yet satisfy this requirement have been converted (e.g. Robotics\cite{ferigo2020gym, aumjaud2021rl_reach}). Moreover, besides theory focused guidelines\cite{henderson2018deep}, practical tools like Garage exist to enable reproducible policy training\cite{garage}.
	
	Nonetheless, RL lacks verifiability of experimental results, such as benchmark submissions. Reviewers shall be enabled to easily verify reported results in figures or tables with minimal effort.
	MDP's already come with the concept of \textit{traces} that are basically a full log of all data (observation, action, reward)\cite{plaat2021drl}, from which figure data could also be constructed (see Section \ref{sec:min-trace}).
	These traces are used for offline reinforcement learning. However, offline RL datasets can become quite large (3TB for the experiments in one paper \cite{agarwal2020optimistic}) and, as a consequence, are hosted on a proprietary central authority.
	To remedy this problem, we collect as little data as possible and without requiring a central hosting authority regardless of size (Section \ref{sec:ipfs}).
	In Figure \ref{fig:rl}, we attempt to visualize the relationship between offline RL datasets, minimal traces, and the data that Garage\cite{garage} collects.
	Thus, we suggest to collect an initial environment configuration corresponding to $s_{init}$ from the Minimal Trace (Section \ref{sec:min-trace}). In Gym, $s_{init}$ contains all values that influence an environments initialization (reset) and transition function (step). Table \ref{table:cartpoleparams} contains an example of $s_{init}$ for CartPole. Moreover, the actions that an agent takes are saved. A consequence is that our method is limited to fully deterministic environments. By properly seeding non-deterministic algorithms' random number generation, non-deterministic problems can also be used. 
	
	We later show that minimal traces allow a compression ratio of up to 77 for regular and up to 12559 for image-based environments (Section \ref{sec:compression}).
	
	Our focus on the environment instead of the policy training is well motivated. The data collected by tools such as Garage\cite{garage} can not overcome one specific problem. The training of neural networks includes certain operations that render it not reproducible if the host machine, host operating system, or software library version change \cite{pytorch-repro}. Unless researchers have the exact same machine and software configuration, reproducibility will become difficult. Software configurations are easily reproducible via Docker\cite{boettiger2015introduction}. However, if the same code produces different results on other machines, then reproducibility becomes practically impossible.
	
	We see minimal traces as an add-on rather than a replacement to current reproducibility approaches.
	It is a workaround because reproducible policy training, such as attempts to Garage\cite{garage} offer, does not yet work. Should training become fully deterministic, minimal traces might become obsolete. Nevertheless, they would still offer the advantage of being computationally cheaper than training, as environment execution requires less computational resources than updating weights of a large neural network. Moreover, environment re-simulation is cheap enough to potentially run in the web browser enabling interactive tools, which is difficult to reproduce with computing clusters needed for training.
	
	Whereas minimal traces are limited to reinforcement learning, their concept transfers perfectly to video games. TrackMania is a great example because its physics is fully deterministic. For its leaderboard, replays are saved. A replay contains the name of a level and all taken actions by the player. Leaderboard submissions are verified for validity\cite{trackmania-replay}. Moreover, it allows to detect Tool-Assisted Runs in pure human play leaderboards\cite{trackmania-tas}.
	More complex games, such as Counter Strike: Global Offensive, Dota 2, and Starcraft, support replays as well. They are also minimal in their sense. However, the multitude of possible inputs is difficult to map directly to a reinforcement learning action space and hence minimal traces. The same applies to the Unreal Engine, which has a general ReplaySystem that can replay any data~\cite{ue4-replay} but does not automatically allow replaying arbitrary scenes deterministically based purely on agent actions. Nevertheless, this larger trove of data is still useful to detect cheats, such as AimBots~\cite{maberry10using}. Furthermore, it can be used to make estimations of player skill~\cite{xenopoulos2020valuing}.
	
	As games are used in competitions, it could also be applied for validation there. For example, two of the games we previously mentioned to support replays have been a competition at the IEEE Conference on Games 2021: Dota 2\cite{font2018dota} and Starcraft\cite{farooq2016starcraft}. Moreover, SpaceInvaders, which we test from Gym, is also a CoG competition\cite{brown2021spaceinvaders}. Other games seem suitable as well, such as GvGAI\cite{gaina20172016}, Snakes\cite{brown2021snakes}, or Bot Bowl\cite{justesen2019blood}. These games might even manage to be directly compatible with minimal traces, as, for example, GvGAI has managed to provide an RL-Gym Environment for its learning track\cite{torrado2018deep}.
	
	Lastly, our data collection suggestion harmonizes with the concept of Procedural Content Generation (PCG), as the seed can be saved for deterministic reproduction. ProcGen has already shown that current RL-Algorithms are struggling to generalize~\cite{cobbe2019procgen}. However, PCG applied correctly already enables better generalization~\cite{risi2020increasing} and more fine-grained training curricula~\cite{green2019evolutionarilycurated},
	and is thus especially important in benchmarking Transfer in Reinforcement Learning~\cite{Muller-Brockhausen21a}.

	\section{Method}
	We will describe our approach in detail, along the lines of 3 different aspects. Minimal traces achieve a high compression ratio compared to regular traces. They can compress up to $10^4:1$ (Section \ref{sec:compression}). Next, we detail how minimal traces enable re-usable visualizations (Section \ref{sec:vis}). Moreover, we suggest the usage of a distributed file system, the interplanetary file system (IPFS)\cite{ipfs}, for long-term storage (Section \ref{sec:ipfs}). 
	Based on these insights, we propose a reproducibility agenda (Section \ref{sec:agenda}).

	\begin{table*}[h]
		\center
		\begin{tabular}{|l|l|l|l|l|l|l|}
			\hline
			Environment & Trace (MB) & Minimal Trace (MB) & Compression Ratio & Training (sec) & Re-simulation (sec) & \% Re-Simulation to Training \\ \hline
			Assault-ram-v0 & 767.21 & 7.76 & 98.82 & 2314.0 & 177.0 & 7.65 \\ \hline
			Assault-v0 & 96165.33 & 7.78 & 12355.67 & 5372.0 & 393.0 & 7.32 \\ \hline
			BipedalWalker-v3 & 530.72 & 181.79 & \textbf{2.90} & 2241.0 & 15.0 & 0.67 \\ \hline
			Breakout-ram-v0 & 757.55 & 8.33 & 90.94 & 2859.0 & 640.0 & \textbf{22.39} \\ \hline
			Breakout-v0 & 96504.98 & 7.98 & 12100.5 & 5647.0 & 520.0 & 9.21 \\ \hline
			CartPole-v0 & 452.25 & 8.5 & 53.23 & 1907.0 & 13.0 & \textbf{0.68} \\ \hline
			Pong-ram-v0 & 767.18 & 7.7 & \textbf{99.58} & 2356.0 & 142.0 & \textbf{6.03} \\ \hline
			Pong-v0 & 94641.99 & 7.54 & \textbf{12559.36} & 5319.0 & 321.0 & 6.03 \\ \hline
			SpaceInvaders-ram-v0 & 767.45 & 7.76 & 98.91 & 2412.0 & 181.0 & 7.5 \\ \hline
			SpaceInvaders-v0 & 95973.03 & 7.75 & 12378.43 & 5448.0 & 378.0 & 6.94 \\ \hline
			Taxi-v3 & 335.85 & 8.46 & 39.69 & 2053.0 & 68.0 & 3.31 \\ \hline
		\end{tabular}
		\caption{A comparison of the space requirements in megabytes and re-simulation cost in seconds for (minimal) traces when training PPO for 1 Million steps and averaged over three runs. We see that minimal traces achieve high compression ratios for single action environments such as Atari or CartPole. Even in the worst case (BipedalWalker-v3), minimal traces still allow a compression ratio of 2.9, saving nearly 350~MB.}
		\label{table:sizes}
	\end{table*}
	
	\subsection{Data compression}
	\label{sec:compression}
	Minimal traces are a way to store experiment signatures efficiently. We will have a closer look at efficient storage for different games. Please refer to Table \ref{table:sizes} for the size comparisons.
	
	Reinforcement learning experiment logs can take up a considerable amount of space. For example, a single work offering an offline RL trace provides 3TB of data \cite{agarwal2020optimistic}.
	In order to compare the amount of space needed for traces vs. minimal traces (Section \ref{sec:min-trace}), we chose different environments with growing observation spaces. Taxi with one integer, CartPole with two floats, BipedalWalker with 24 floats, Atari-Games RAM using 128 bytes, and the produced 210 x 160 pixel RGB image using 100800 bytes. More interestingly, whereas most environments have only a single integer action space $A$, BipedalWalker has three continuous actions increasing the space required for minimal traces.
	We train on the environment for 1 Million steps using PPO\cite{schulman2017proximal} from stable baselines 3\cite{stable-baselines3}, using the default hyperparameters for both environment and agent.
	During training, we collect the full MDP-trace (Observation, Action, and Reward) and the minimal MDP-trace (Env-Params and Actions) for each environment.
	The results in Table \ref{table:sizes} are striking: Minimal traces enable a compression ratio of 12559.36 for image-based environments with a single action, such as Atari Pong. For the 128-byte ram observation, the compression ratio falls to 99.58 for Pong, reducing 767.18~MB to 7.7~MB.
	
	Nevertheless, environments with small observation spaces such as CartPole still allow a compression rate of 53, reducing a 452.25~MB trace down to 8.5~MB.
	However, BipedalWalker underlines that the potentially saved space depends solely on the size difference between the observation space $S$ and the action space $A$. For BipedalWalker, 24 numbers in the observation space vs. 4 in the action space. Hence the compression ratio of 2.9 reduces the 530.72~MB trace down to 181.79~MB. An intriguing discovery we made is a varying size for the re-simulated trace of BipedalWalker, which should not occur. Thus we performed an experimental analysis and found  that 5 in 100 re-simulations yielded a different trace due to rounding errors. Consequently,  BipedalWalker is not yet fully deterministic. We repeated this experiment for all other tested environments in Table 1 and found that they are fully deterministic, hence yielding 0 failed re-simulations.

	We also measured the time to re-simulate a full MDP-trace from a minimal MDP-trace vs. the training time. Our measurements are also shown in Table \ref{table:cartpoleparams}. Note that for timing-related data, there is variation in runs for different hardware. All experiments were run on a machine with an Intel Xeon Silver 4214, an Nvidia Geforce RTX 3090, and 256GB of RAM.
	The cost of re-simulation depends on the complexity and observation space of the environment. 
	For Atari, re-simulation varies per observation and game. In the worst case, it takes 22.39\% of the training time for Breakout-ram-v0, and in the best case 6.03\% for Pong-v0.
	Computationally less intensive environments, such as BipedalWalker-v3 or CartPole can lower this further to less than 1 \% (0.68\%) of training time. The main reason for re-simulation outperforming training is that many episode traces can be re-simulated in parallel.
	
	\subsection{Re-Usable Visualizations}
	\label{sec:vis}
\begin{figure}
		\centering
		\includegraphics[width=0.45\textwidth]{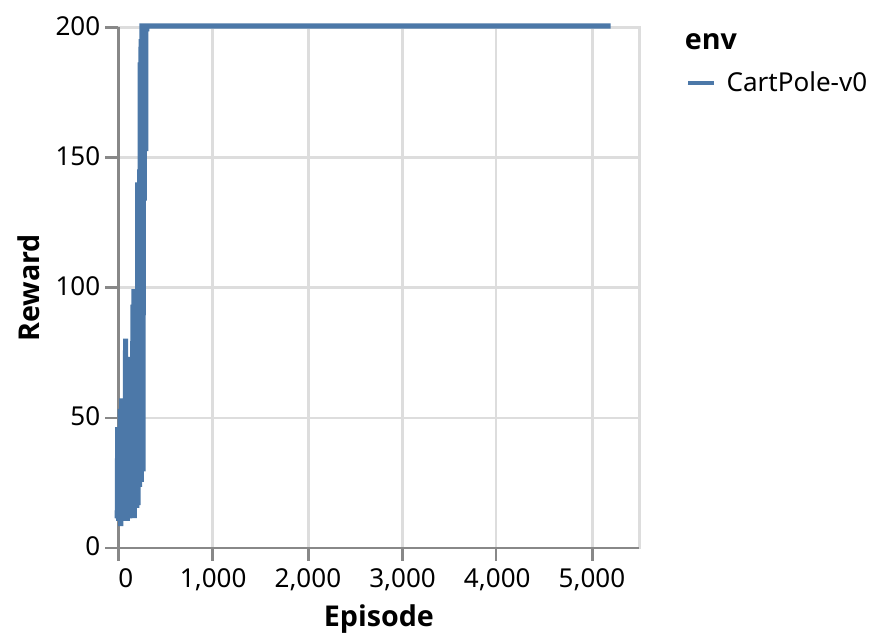}
		\caption{A re-usable Vega Lite Reward Graph, generated from the Description shown in Listing \ref{lst:graph}. It displays the sum of reward per episode achieved during training of PPO3\cite{schulman2017proximal} for 1 million steps using the default hyperparameters for algorithm and environment. The used reward data can be re-simulated based on the minimal trace with ID "2IEpetGGwNFOs38rhAFHx".}
		\label{fig:graph}
	\end{figure}

	\begin{lstlisting}[
		caption=An excerpt for a re-usable Vega Reward Graph description. Generates the graph shown in Figure \ref{fig:graph}. The full JSON-File that can be explored in the Vega-Editor can be found in the Code-Repository as listing\_graph.json.,
		label=lst:graph,
		frame=single
		]
{"$schema": "...",
  "description": "Reward per Episode",
  "data": {"values": [{"Episode": 0,
    "Reward": 11.0, "env": "..."}, ...]},
  "mark": "line",
  "encoding": {
    "x": {"field": "Episode",
      "type": "quantitative"},
    "y": {"field": "Reward",
      "type": "quantitative"},
    "color": {"field": "env",
      "type": "nominal"}
  }
}
	\end{lstlisting}
	Khetarpal et al.~\cite{khetarpal2018re} suggest saving the values that are used to plot figures and providing the code to the plot. 
	This improves reproducibility when the numbers are the results of the experiments. Minimal traces enable re-simulating this data to verify if this is the case.
	Moreover, minimal traces allow looking at different values than presented in a paper. For example, if a paper reported only the average reward, one could extract the median reward instead through the re-simulated data. Alternatively, if reward per episode was reported, one could instead look at reward per step.
	To increase the re-usability and accessibility of figures, we suggest using Vega\cite{vega}.
	Vega is a JSON-based graph description language. The main advantage is that plotted values are embedded inside the human-readable JSON data. Hence, one could extract a baseline algorithm reward line from the Vega description of an original paper and then compare it to a newly trained variation without re-simulation. Of course, it still allows exporting a scalable graphic for usage inside of the paper (e.g., Figure \ref{fig:graph}).
	In this case, guidelines on designing a proper baseline\cite{islam2017reproducibility} are not that important anymore because the actual data from other papers can be directly compared\footnote{Assuming they are being compared on the same benchmark environment with the same configuration}.
	
	Another advantage of Vega\cite{vega} is the ability to load Graph-Definition-Files in a Browser. These come with various advantages, such as seeing the exact value of an individual coordinate, zooming, scrolling through the axes, and changing colors if these are not color-blind friendly.
	
	\subsection{Data availability}
	\label{sec:ipfs}
	The data we suggest to collect (Section \ref{sec:method}) potentially requires much storage. Whereas some hosting authorities allow researchers to upload large datasets for long-term availability, the two main problems with a central authority are the possibility that data could be tampered with / changed and that there is only a single point of failure regarding availability.
	The Interplanetary File System\cite{ipfs} lets us address these issues. It behaves similar to BitTorrent. Users who downloaded a file also share it, eliminating the need for a central server to store all files.
	Moreover, the data is hashed, so it can not be altered afterward by the original author, the hosting authority, or a redistributing user. A hash can represent a full folder with many subfiles that also all have individual hashes / IDs.
	So the log file itself and the results produced with it become verifiable.
	IPFS \cite{ipfs} also advertises its usefulness for scientific purposes, and we believe our minimal traces are well suited for it.
	Moreover, IPFS can co-exist and even integrate into a hosting service (such as Zenodo\cite{zenodo}) that could mirror the data it provides via IPFS. That would allow them to use it as a Content Delivery Network (CDN) for the actual files listed in the Zenodo database.
	Finally, conferences or journals could have public lists of relevant dataset IDs for published papers that should be \textit{pinned}. Pinning in IPFS can be seen as the equivalent of mirroring data. A pinned ID will permanently (until unpinned) be held in local storage and thus be available to others requesting it.
	
	\section{Reproducibility Agenda}
	\label{sec:agenda}
	Minimal traces are a useful step on the path to reproducibility, allowing efficient verification of experiments. To further improve the reproducibility of the field and to put our work in perspective, we suggest the following agenda.
	
	We propose researchers experimenting on deterministic Environments (ideally using PCG in order to improve generalizability) to do the following:
	\begin{enumerate}
		\item{Use the verifiability tool to collect minimal traces}
		\item{Provide source code, including a runnable container (e.g. Docker), to allow verification of results and figures}
		\item{Generate figures using a common visualization grammar such as Vega\cite{vega}, facilitating re-use of figure data}
		\item{Utilize IPFS\cite{ipfs} to ensure the data's integrity and long-term availability}
	\end{enumerate}
	
	\subsection{Agenda applied to CartPole}
	\label{sec:apply}	
	To better illustrate our suggested agenda in practice, we will illustrate how following it looks for the proverbial CartPole environment.
	In Table \ref{table:cartpoleparams}, we have prepared the environment hyperparameters ($s_{init}$) relevant for each CartPole episode. Garage\cite{garage} (Figure \ref{fig:rl}) also collects these parameters but assumes policy training to be fully reproducible, which it is currently not (Section \ref{sec:method}).	
	\begin{table*}[]
		\center
		\begin{tabular}{|l|l|l|}
			\hline
			\textbf{Name}           & \textbf{Default}        & \textbf{Description}                                                       \\ \hline
			Gravity                 & 9.8                     & Gravitational Power of the Planet                                          \\ \hline
			Pole Mass               & 0.1                     & The mass of the pole balancing on the cart                                 \\ \hline
			Pole Length             & 0.5                     & The length of the pole (The actual length is this value times two)         \\ \hline
			Cart Mass               & 1                       & The mass of the cart                                                       \\ \hline
			Force Magnitude         & 10                      & Strength of the force applied to the cart on input                         \\ \hline
			Tau                     & 0.02                    & How much time passes between state updates (in seconds)                    \\ \hline
			Theta Threshold Radians & $\sim$0.209 ($\sim$12°) & The angle at which the pole is considered to be tipped over                \\ \hline
			X Threshold             & 2.4                     & The area in which the cart may move without being considered out of bounds \\ \hline
			Use euler kinematics    & TRUE                    & Influences cart movement                                                   \\ \hline
			Random Seed             & varies                  & Influences initial cart position and pole angle                            \\ \hline
		\end{tabular}
		\caption{Complete list of initial Environment Parameters $s_{init}$ (Section \ref{sec:min-trace}) for Cartpole.\\Given the same $s_{init}$ and action sequence $\alpha$, a deterministic MDP like CartPole will always yield the same trace.}
		\label{table:cartpoleparams}
	\end{table*}
	In Listing \ref{lst:code}, we show that recording minimal traces merely requires wrapping the Gym-environment. The code in this listing is not pseudo-code but the actual main file of our runnable example. The last function generates a Vega\cite{vega} (Section \ref{sec:vis}) JSON-Description (Listing \ref{lst:graph}) that can be used to visualize a graph (Figure \ref{fig:graph}). It can also be pasted into the Vega-Editor\footnote{https://vega.github.io/editor/}.
	Table \ref{table:sizes} is also based on re-simulated data from minimal traces.

	New projects wanting to record minimal traces need to wrap their Gym-environment with the \textit{record} function from the \textit{vgym}-folder\cite{code} before training. Then, the minimal traces will be saved into a sub-folder of the current working directory. A minimal trace file is serialized into the Concise Binary Object Format (CBOR)\cite{bormann2013concise} and compressed with zlib\cite{deutsch1996zlib}.
	Note that minimal trace sizes in Table \ref{table:sizes} reflects the size before serialization and compression.
	The utility function \textit{load\_replay} in our package abstracts these serialization details away from the user.
	After loading a file, either all episodes can be re-simulated into regular traces in parallel using \textit{resimulate\_parallel}, or an individual episode can be re-simulated using \textit{episode\_to\_trace}.
	Based on that, re-simulated data figures and tables should be generated, and the code as well as recorded minimal traces published alongside the submission.
	
	\begin{lstlisting}[
		caption=Example code that trains a policy on Cartpole collecting a minimal trace. The full trace is re-simulated based on the minimal trace and used to generate a Vega Graph-Description.,
		label=lst:code,
		frame=single
		]
import gym
from vgym import record, resimulate
from stable_baselines3 import PPO
from example import make_graphs_from
env = gym.make("CartPole-v0")
renv = record(env, name="CartPole-v0")
renv.reset()
PPO("MlpPolicy", renv, verbose=1)
  .learn(1000000)
# Recreate Trace from minimal Log
trace = resimulate(renv,
  renv.minimal_traces.to_list())
# Plot using re-simulated data
make_graphs_from([("CartPole-v0", trace)])
		
	\end{lstlisting}
	Our code repository contains a Dockerfile, recorded / used minimal traces, the example code behind the imported functions of Listing \ref{lst:code}, and a readme detailing how to re-simulate the shown graph (Figure \ref{fig:graph}), as well as the values for the size comparison table (Table \ref{table:sizes}).
	All data is hosted on IPFS\cite{ipfs} here\cite{code}.
	
	\section{Discussion}
	If properly applied, the concept of minimal traces allows for reproducible and verifiable reinforcement learning experiments. Moreover, it enables re-usability of result data, more accessible ways to view the data interactively, inspection of individual episodes and data-saving for offline RL datasets.
	
	A limitation of minimal traces is that the source of the action sequence can not properly be verified.
	From the obtained data alone one cannot exclude any sophisticated ways of cheating the authors may have applied.
	Whereas a result figure or table may be fully reproduced with our data, one can not know whether the agent created the supplied action sequences. The data could be handcrafted or made by a heuristic or any other method that is not listed in the reviewed paper.
	Although minimal traces do not yet achieve end-to-end verifiable research experiments, they are an important step towards verifiable experiment figures to show that values portrayed in figures or tables have truly been achieved within the tested environment and have not been randomly generated.
	In combination with host executed competition benchmarks such as GVGAI, they enable post hoc analysis of individual agent performances.
	
	\subsection{Environmental Impact}
	In times where big research experiments use large amounts of electricity\cite{patterson2021carbon}, it is important to note the possible environmental impact of our suggestion.
	Hard-disk space seems cheaper than the high wattages of training an agent on a GPU. Hard-disk space instead of computation is also used to \enquote{greenwash} novel blockchains as Chia\cite{cohen2019chia}. Nevertheless, nothing electronic comes free, so the hardware still needs to be built, and energy needs to be used to power the servers pertaining the datasets we suggest to collect. In consequence, it will increase the global environmental impact of RL. However, we see no other, more minimal, and less intrusive way to ensure full verifiability of RL research. Thanks to the suggestion to share the data via IPFS\cite{ipfs}, the servers would not need to stay online 24 / 7. There could be specific times of data availability where the server is switched on and the data accessible while ensuring the data is not tampered with. Moreover, the distributed nature of IPFS might make the necessity of a central storage server obsolete, given enough participants pertaining parts of the datasets.
	
	\section{Conclusion}
	\label{sec:conclusion}
	In reinforcement learning, reproducibility of experimental results, and verification of research claims, is an important challenge. 
	Our work introduces a methodology to verify experimental results building on the concept of minimal traces. We provide a full implementation of this method and have tested it on small and larger reinforcement learning experiments. 
	
	For typical experiments, minimal traces enable compression ratios of up to 12539, reducing an offline RL trace of Atari Pong from 94~GB down to 8~MB. Moreover, re-simulating this minimal trace back to its original size takes 6\% of the original training time.
	
	As our example shows, the collection of minimal traces requires but a wrapper around a Gym-environment.
	
\begin{figure}[h]
		\centering
		\includegraphics[width=0.5\textwidth]{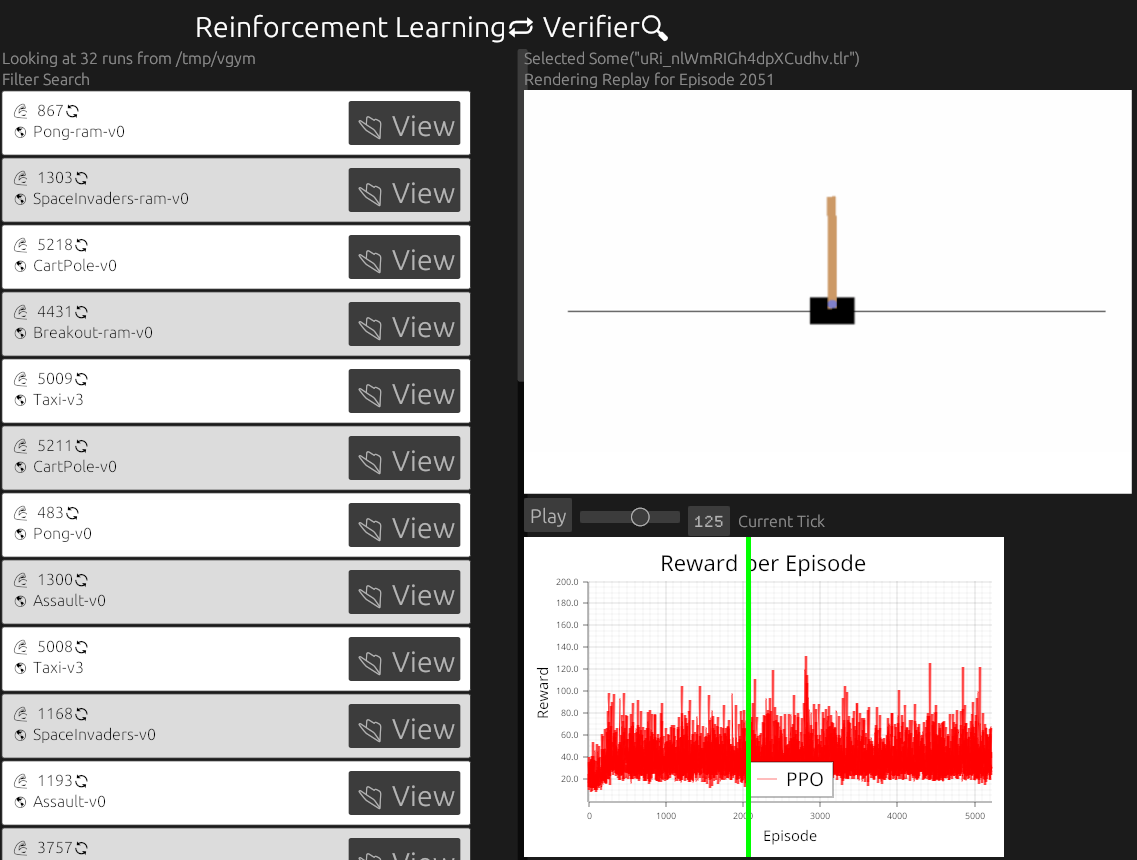}
		\caption{Mock-up of a web-based tool we envision for future work. The left side contains a list of minimal traces, the environment, and amount of contained episodes. The right side shows details of the re-simulated trace. For inspection, one can select individual episode replays directly from the training reward graph, highlighted by the green vertical stripe. Each episode can be stepped through frame by frame.}
		\label{fig:mockup}
	\end{figure}

	While minimal traces are limited to deterministic Reinforcement Learning problems, the idea transfers well to (video) games. Trackmania already applies leaderboard verification via replays, showing that benchmarks and competitions could adopt similar concepts.
	We envision a web-based tool that allows re-simulation and verification of results without any software setup by a reviewer on their local machine for future work. To that end, we provide a mock-up (Figure  \ref{fig:mockup}) with a functionality description. 
	This could be implemented through either a Rust-Gym port of the current approach or by having a Python interpreter that properly works in a web assembly environment and with Gym.

	\bibliographystyle{IEEEtran}
	\bibliography{paper}
	
\end{document}